\documentclass[10pt,twocolumn,letterpaper]{article}

\usepackage{iccv}
\usepackage{times}
\usepackage{epsfig}
\usepackage{graphicx}
\usepackage{amsmath}
\usepackage{amssymb}

\usepackage[pagebackref=true,breaklinks=true,letterpaper=true,colorlinks,bookmarks=false]{hyperref}
\usepackage{xurl}

\usepackage{booktabs}
\usepackage{pifont}
\usepackage{xcolor,colortbl}
\usepackage{float}
\definecolor{darkergreen}{RGB}{21, 152, 56}
\definecolor{red2}{RGB}{252, 54, 65}

\newcommand{\LL}{\mathcal{L}}

\definecolor{mygray}{gray}{.88}
\definecolor{mygraylite}{gray}{.93}

\iccvfinalcopy 

\def\iccvPaperID{****} 

\begin{document}


\title{MEAL V2: Boosting Vanilla ResNet-50 to 80\%+ Top-1 Accuracy \\on ImageNet without Tricks}

\author{Zhiqiang Shen~~~~~~~~~~~~~~ Marios Savvides\\
Carnegie Mellon University\\
{\tt\small Code:~\url{https://github.com/szq0214/MEAL-V2}}
}

\maketitle

\begin{abstract}
We introduce a simple yet effective distillation framework that is able to boost the vanilla ResNet-50 to 80\%+ Top-1 accuracy on ImageNet without tricks. 
We construct such a framework through analyzing the problems in existing classification system and simplify the base method {ensemble knowledge distillation via discriminators}~\cite{shen2019meal} by: (1) adopting the similarity loss and discriminator only on the final outputs and (2) using the average of softmax probabilities from all teacher ensembles as the stronger supervision. Intriguingly, three novel perspectives are presented for distillation: (1) weight decay can be weakened or even completely removed since the soft label also has a regularization effect; (2) using a good initialization for students is critical; and (3) one-hot/hard label is not necessary in the distillation process if the weights are well initialized. We show that such a straight-forward framework can achieve state-of-the-art results without involving any commonly-used techniques, such as architecture modification; outside training data beyond ImageNet; autoaug/randaug; cosine learning rate; mixup/cutmix training; label smoothing; etc. 
Our method obtains {\bf \em 80.67\%} top-1 accuracy on ImageNet using a single crop-size of 224$\times$224 with vanilla ResNet-50, outperforming the previous state-of-the-arts by a significant margin under the same network structure. Our result can be regarded as a strong baseline using knowledge distillation, and to our best knowledge, this is also the first method that is able to boost vanilla ResNet-50 to surpass 80\% on ImageNet without architecture modification or additional training data. On smaller ResNet-18, our distillation framework consistently improves from 69.76\% to {\bf \em 73.19\%}, which shows tremendous practical values in real-world applications. 
\end{abstract}

\section{Introduction}

\begin{figure}[t]
  \centering
  \includegraphics[width=0.36\textwidth]{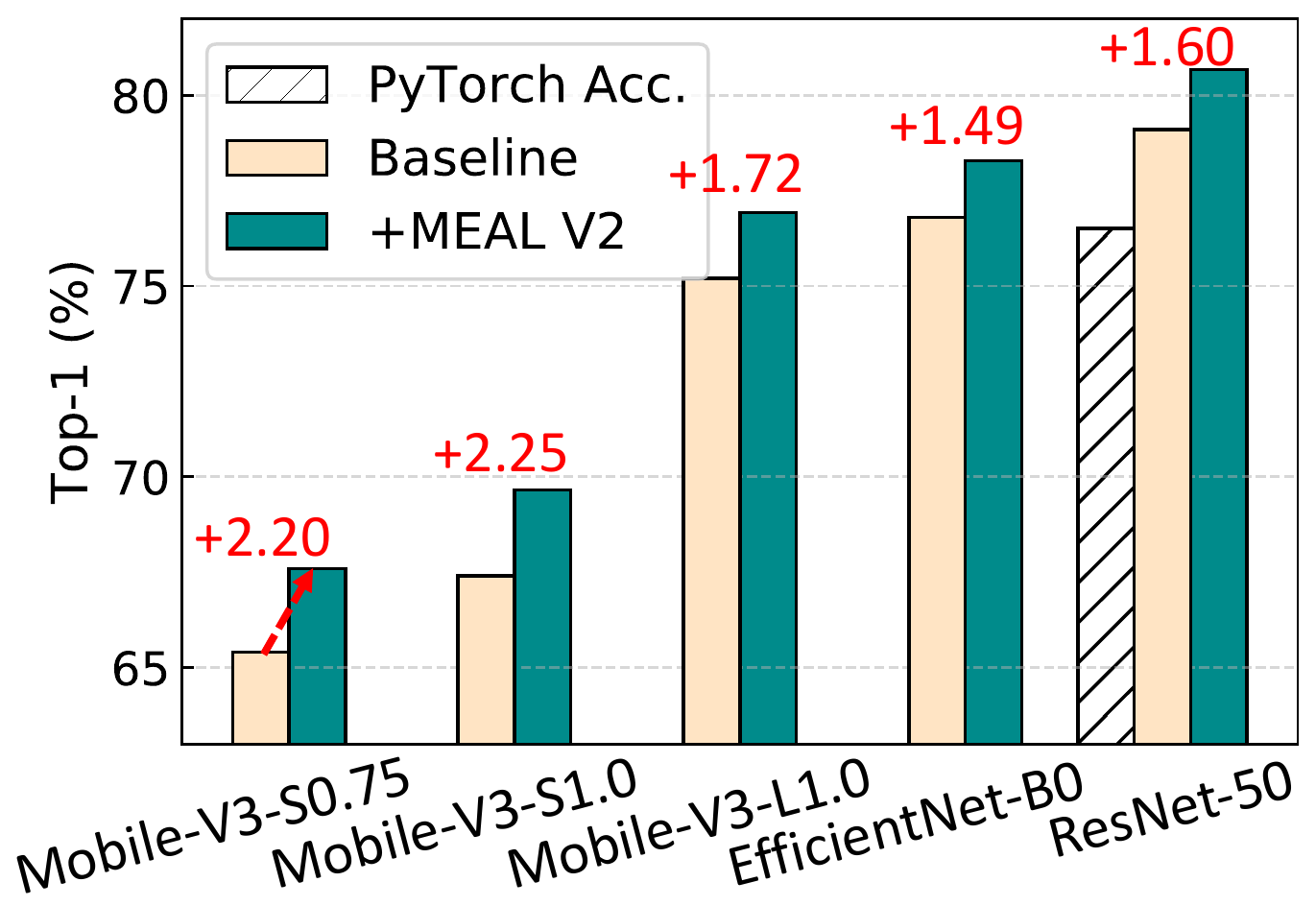}
  \vspace{-0.05in}
  \caption{An illustration of improvement with our MEAL V2 on ImageNet. The architectures from left to right are MobileNet V3-Small 0.75~\cite{howard2019searching}, MobileNet V3-Small 1.0~\cite{howard2019searching}, MobileNet V3-Large 1.0~\cite{howard2019searching}, EfficientNet-B0~\cite{tan2019efficientnet} and ResNet-50~\cite{he2016deep}.}
  \label{fig:improvements}
  \vspace{-0.05in}
\end{figure}

Convolutional Neural Networks (CNNs)~\cite{lecun1998gradient} have been proven useful in many visual tasks, such as image classification~\cite{krizhevsky2012imagenet,he2016deep}, object detection~\cite{girshick2014rich,ren2015faster}, semantic segmentation~\cite{long2015fully}, as well as some particular scenarios, like transferring feature representation~\cite{yosinski2014transferable}, learning detectors from scratch~\cite{shen2017dsod}, etc. In order to achieve highest possible accuracy, many training techniques and data augmentation methods have been proposed, such as mixup~\cite{zhang2017mixup}, cutmix~\cite{yun2019cutmix}, autoaug~\cite{cubuk2018autoaugment}, randaug~\cite{cubuk2020randaugment}, fix resolution discrepancy~\cite{touvron2019fixing}, etc. Some works also focus on modifying the network structures, e.g., SE module~\cite{hu2018squeeze}, Stem block~\cite{shen2017dsod}, Split-attention~\cite{zhang2020resnest}. This paper is to similarly obtain the best possible performance of a network, but our proposed method is orthogonal to the above techniques. In general, our method only relies on a {\em teacher-student} paradigm with a powerful ensemble of teachers and a good initialization of the student. It is simple, straight-forward, but effective and can achieve state-of-the-art performance on large-scale dataset. The advantages of our method are: 1) no architecture modification; 2) no outside training data beyond ImageNet; 3) no complex learning rate scheduler like cosine {\em lr}; 4) no extra data augmentation like mixup, autoaug, etc.

The objective of this paper is to give a better understanding on knowledge distillation and promote the capability and robustness of the classification networks through distillation. 
We first analyze and introduce several critical factors and limitations that will degrade the performance in the existing classification systems. We find the main drawback in the conventional training strategy of a network, i.e. with the one-hot label, is the inferior ability to distinguish the semantically similar categories, as shown in Fig.~\ref{fig:motivation} (1). Networks trained with one-hot label are incapable of handling the semantically similar instances. We observe knowledge distillation~\cite{hinton2015distilling} is surprisingly effective in dealing with this circumstance as the supervision from teacher networks is smoothed and much lower than one-hot value. Therefore, the distilled students will encourage representations of examples to lie in tight equally separated clusters and enforce similar instances more distinguishable in feature space, similar to label smoothing~\cite{muller2019does,shen2021is}. We show multiple promising improvements on various network architectures using distillation in Fig.~\ref{fig:improvements}. The potential improvement of our method can be larger if replacing with stronger teachers.

We also have a few interesting discoveries in our training process, for example, among them we would like to {emphasize that the one-hot/hard label is not necessary if the weights are already well initialized and could not be used in the distillation process~\cite{hinton2015distilling,shen2019meal}. Some discussions about this perspective are provided in our Appendix. Also, weight decay can be weakened or even removed since the soft label also has a regularization effect, and using a good initialization is critical for distillation. 
While some previous studies deem that structure might be more crucial than pre-trained parameters on some downstream tasks like object detection~\cite{shen2017dsod}, segmentation~\cite{jegou2017one}, etc., we still believe that boosting the performance of standard and classical network structures is interesting and useful, especially if the networks are already small and compact, like MobileNet V3, EfficientNet-B0, since the proposed method can be effortlessly generalized to other elaborated or searched architectures. 
Our method can be considered as a post-process to distill small and compact models for further boosting their performance, while no modification is required.

Our contributions in this paper are as follows:

\vspace{-0.12in}
\begin{itemize}
	\addtolength{\itemsep}{-0.1in}
	\item We provide empirical analysis and insights through individual class's accuracy to expose the mechanism of how knowledge distillation helps classification. It is not trivial to understand the principles behind it.
	\item We present a simple yet effective and practical framework that can boost the performance of existing tiny models by a significant marginal. We show evidence and provide a demonstration, and also give detailed guidance on how to establish such a strong framework.
	\item We further visualize our model to explore where the superior performance is from. Moreover, we transfer trained parameters to other datasets like fine-grained recognition to show the transferability of our method.
\end{itemize}

\section{Related Work}

\noindent{\textbf{Image Classification.}}
Image classification is a fundamental task in computer vision. AlexNet~\cite{krizhevsky2012imagenet} is considered as the seminal design that is proven feasible for deep neural networks on the large-scale datasets. After that, many innovative network structures have been proposed. Szegedy et al.~\cite{szegedy2015going} proposes an ``Inception'' design that concatenates features maps produced by various sizes of kernels. He et al.~\cite{he2016deep} creatively proposed residual blocks with skip connections, which is firstly enable to train extremely deep networks more than 100 layers. Huang et al.~\cite{huang2017densely} further proposed densely layer-wise connections for building DenseNet. Besides, some architectures are also targeting at mobile device scenario, such as MobileNet series~\cite{howard2017mobilenets,sandler2018mobilenetv2,howard2019searching}, ShuffleNet~\cite{zhang2018shufflenet,ma2018shufflenet}, etc. 
With the development of these modern neural network designs and automatic architecture search~\cite{zoph2016neural,tan2019efficientnet}, this task has been one of the fastest moving areas and achieved surprising results which even surpasses human-level performance on large-scale datasets like ImageNet~\cite{deng2009imagenet} and OpenImage~\cite{kuznetsova2018open}.

\noindent{\textbf{Knowledge Distillation.}}
Hinton et al.~\cite{hinton2015distilling} pioneered the concept of distilling knowledge from a larger teacher network or ensemble into a smaller compressed student. Mathematically, this paradigm of training the student on softened teacher predictive distribution is using the conventional cross-entropy with predicted labels. The student is encouraged to mimic the teacher output distribution, which helps the student generalize much better on validation set and in certain cases leads to the student performing even better than the teacher itself. These studies argued that the teacher distribution provided much richer information about an image compared to just one-hot labels. Further studies extended this concept by using internal feature representations~\cite{romero2014fitnets,shen2019meal}, adversarial training with discriminators~\cite{shen2019meal} and transfer flow of solution procedure matrix as the student initialization~\cite{yim2017gift}. Some works also proposed online distillation~\cite{walawalkaronline,zhu2018knowledge} that do not rely on a pre-trained teacher, so teacher and student can be learned simultaneously.

\noindent{\textbf{Network Compression.}} Knowledge distillation is a natural way to produce the compressed student network through imitating the teacher's softened prediction. Besides it, other methods like weight quantization~\cite{hubara2017quantized,zhu2016trained,jacob2018quantization} and binarization~\cite{courbariaux2016binarized,rastegari2016xnor}, weight pruning~\cite{han2015learning,han2015deep} and channel pruning~\cite{li2016pruning,liu2017learning,he2017channel} can also achieve the compression  purpose. Knowledge distillation differs from them as the compressed network is designed before training so there is no additional operation required, such as reconstruction, retraining, etc.

\begin{figure*}[t]
  \centering
     \hspace{0.1in}
  \includegraphics[width=0.90\textwidth]{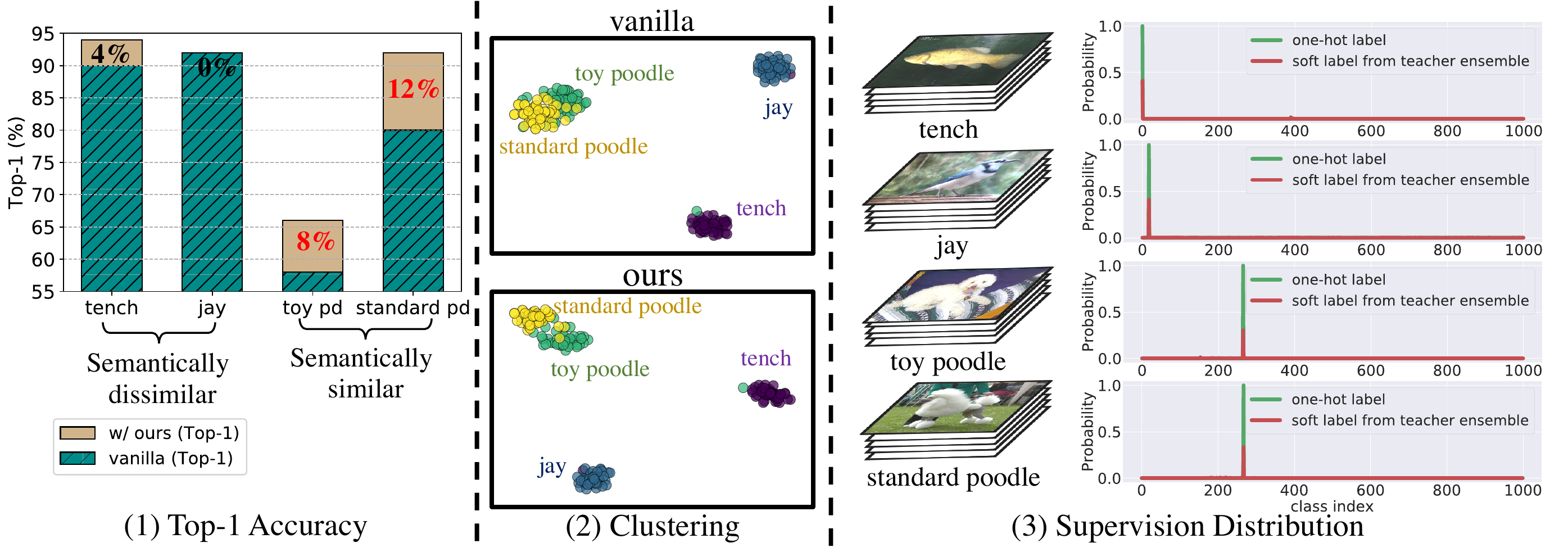}
  \vspace{-0.08in}
  \caption{(1) is the comparison of class-wise accuracy on four semantically similar and dissimilar classes between PyTorch official model and ours. (2) is the feature embedding visualization using t-SNE~\cite{maaten2008visualizing} on these four classes. (3) is the visualization of supervision for one-hot label and soft label in distillation.}
  \label{fig:motivation}
  \vspace{-0.15in}
\end{figure*}

\section{Analysis: Problems in Existing Classification System}

Consider a classification task that distinguishes various breeds of dogs, e.g. toy poodle, miniature poodle, etc., the output predictive distribution of a higher capability teacher always provides the student model with the extra information of how alike one breed of dogs looks to the other. This helps the student learn more generalized features of each dog breed compared of providing just one hot-label which fails to supply any comparative knowledge. Also, in the process of trying to mimic the distribution of a much deeper teacher model the student tries to find a compact solution of transformation. This inherently enforces the student to explore more informative knowledge and generalize better. Impressively, in certain cases the student manages to outperform its teacher due to this superior generalization.

Training with one-hot labels accompanying with cross-entropy loss is a ``balanced'' learning system, which means the objective will enforce each class to be equidistant to all remaining class’s distance, so the learned model is not sensitive to the semantically similar classes (e.g., different dog breeds) or dissimilar classes (e.g., dog and fish). Some situations it even gives risk to performing with respect to incorrect annotation of classes. By using dynamical soft labels from knowledge distillation of a strong teacher, different examples from the same or different classes can have very different similarities to other classes, thus the student can capture additional subtle information and prevent from overfitting. We illustrate training curves in Fig.~\ref{fig:overfitting} by using one-hot/hard and soft labels. We use exactly the same training hyper-parameters and settings, good initialized parameters, learning rate schedule, etc. We found if the initialization is already well-learned, training with one-hot label is easy to be overfitting (blue curves), while with soft labels (orange curves), the model can continue to learn new knowledge and generalize better on the validation set. Moreover, we select two semantically similar classes (toy  poodle and standard poodle) and two dissimilar classes (tench and jay), and test their accuracy on ImageNet val set with the moderate and extremely good ResNet-50\footnote{PyTorch official model (76.15\%) and ours (80.67\%).}. Results are shown in Table~\ref{im:gap}, interestingly, it can be observed the overall accuracy gap mainly hinges on the semantically similar classes. Intuitively, these classes are difficult to distinguish in a classification system thus the bottleneck also lies here. 
In this paper, we aim to diagnose these problems, we will give insights about how distillation method alleviates them below.

\begin{figure}[t]
  \centering
  \includegraphics[width=0.48\textwidth]{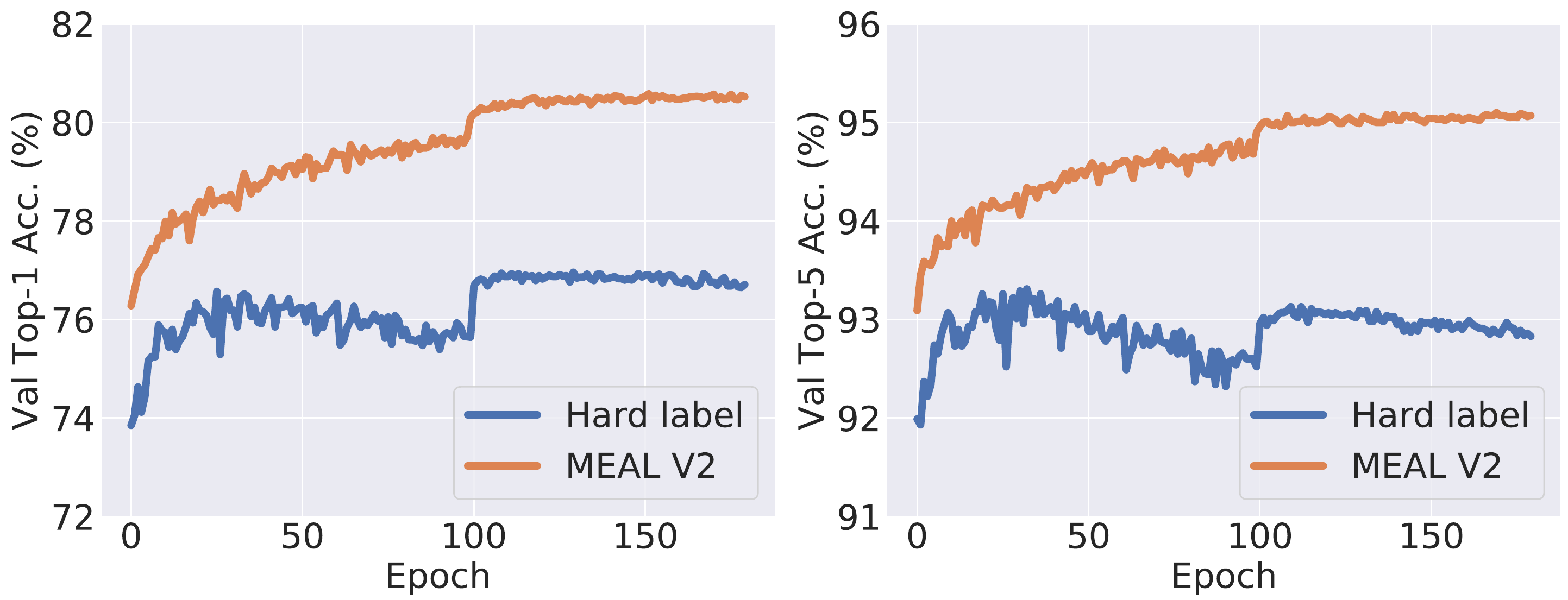}
  \vspace{-0.2in}
  \caption{Comparison of Top-1 (left) and 5 (right) val accuracy by using hard and soft labels. We use exactly the same training hyper-parameters and settings, including the same initialized parameters, learning rate schedule, etc., but the different supervision shapes.}
  \label{fig:overfitting}
  \vspace{-0.05in}
\end{figure}

{
\renewcommand{\arraystretch}{1.2}
\setlength{\tabcolsep}{0.2em}
\begin{table}[t]
  \caption{Comparison of class-wise accuracy on four semantically similar and dissimilar classes. {``Vanilla''} is PyTorch official model.}
  \label{im:gap}
\resizebox{0.48\textwidth}{!}{
\begin{tabular}{c|c|c||c|c|c}
\toprule[1.1pt]
        & \multicolumn{2}{c||}{\bf Semantically dissimilar} & \multicolumn{2}{c|}{\bf Semantically similar} &  Accuracy     \\ \hline
Model   & tench (\%)           & jay (\%)         & toy  poodle (\%) & standard poodle (\%)  & \bf Avg (\%)  \\ \hline
Vanilla & 90                     & 92                  & 58           & 80            & 76.15 \\ \hline
\bf Ours    &\bf  94$^{\bf +4\%}$   & \bf 92$^{\bf +0\%}$     & \bf 66 $^{\bf +8\%}$  & \bf 92$^{\bf +12\%}$  & \bf 80.67 $^{\bf +4.52\%}$ \\ 
\bottomrule[1.1pt] 
\end{tabular}
}
\vspace{-0.1in}
\end{table}
}

\section{Solution: What Can KD Solve?}
\vspace{-0.05in}
In this section, we discuss the following aspects that knowledge distillation can handle in the modern classification system: 
(1) enlarge the distance of samples between semantically similar classes; (2) overcome multiple objects problem; (3)  take advantage of random crop augmentation and avoid its limitations. 

\subsection{Semantically Similar and Dissimilar Classes}

As aforementioned in Table~\ref{im:gap} and Fig.~\ref{fig:motivation} (1), the performance of good or moderate models primarily depends on the semantically similar classes. To further diagnose how this happens, we visualize the embedding distributions in Fig.~\ref{fig:motivation} (2). As expected, the clusters of toy and standard poodle breeds are mixed up together, and tench and jay are separate from each other. Impressively, the model trained with knowledge distillation can split the similar classes representations to some extend and make the border of two clusters clearer, which greatly facilitates the final classifying. Let us take this one step further to see how knowledge distillation obtains this function, we visualize the supervisions from teacher ensemble in Fig.~\ref{fig:motivation} (3). This is the main difference in our framework when learning with one-hot or distilled supervisions. we are interested and illustrate the supervisions from the training set since those are the ones used for distillation. We show the averaged probability of all samples in each class. The prediction of major class is only 0.3$\sim$0.4 comparing to 1.0 as one-hot labels. These softened labels make models less confident and generalize much better, especially on those semantically similar classes.

\begin{figure}[t]
  \centering
  \includegraphics[width=0.4\textwidth]{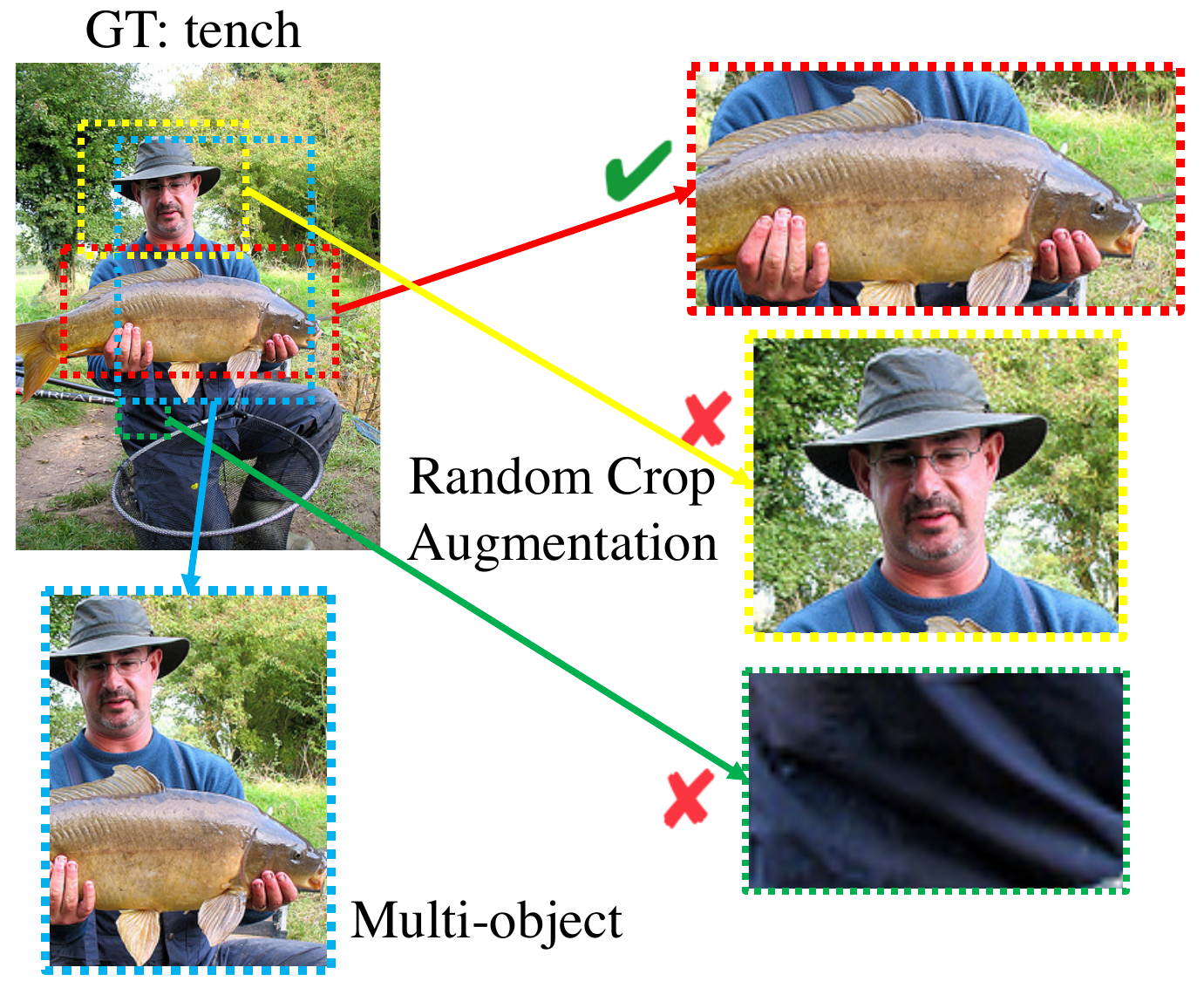}
  \caption{Illustration of random crop data augmentation strategy on an image. This strategy randomly crops regions crossing a predefined scope, e.g. 8\%$\sim$100\% of the whole image size on ImageNet but it will involve massive noises and incorrect labels.}
  \label{fig:random_crop}
  \vspace{-0.05in}
\end{figure}

\subsection{Multi-Object/Random-Crop Issues}

Multi-object is the scenario where there are at least two categories in which the images can be classified, this is a widely existing situation in ImageNet dataset as shown in Fig.~\ref{fig:random_crop}. The one-hot label of this image is {\em tench} but a {\em human} is also contained in it and the area is even larger than the {\em tench}. It causes a label mismatch problem if using standard one-hot label for training. This paper argues that distillation can tackle this circumstance as the label is predicted by a well pretrained teacher network and label distribution depends on the content of input image instead of the assigned labels. Also, the soft label can be a multi-peak distribution which can model the mixed information of multiple objects. 

Random crop data augmentation is an indispensable technique heavily utilized in modern network training. While as shown in Fig.~\ref{fig:random_crop}, this strategy tends to involve a large proportion of noise if cropping on the background area or a small region of objects, which means it will always result in inaccurate labels for the augmented regions. 
In the case of incorrect region by the random cropping, the global ground-truth label is used for such input. Despite deep neural networks are highly tolerant to noises in labels but the incorrectness will inevitably impair the effectiveness of learning process. Instead, knowledge distillation will predict the true probability distribution for each input independently and reflect what the input patch really is.

\begin{figure*}[t]
  \centering
     \hspace{0.1in}
  \includegraphics[width=0.88\textwidth]{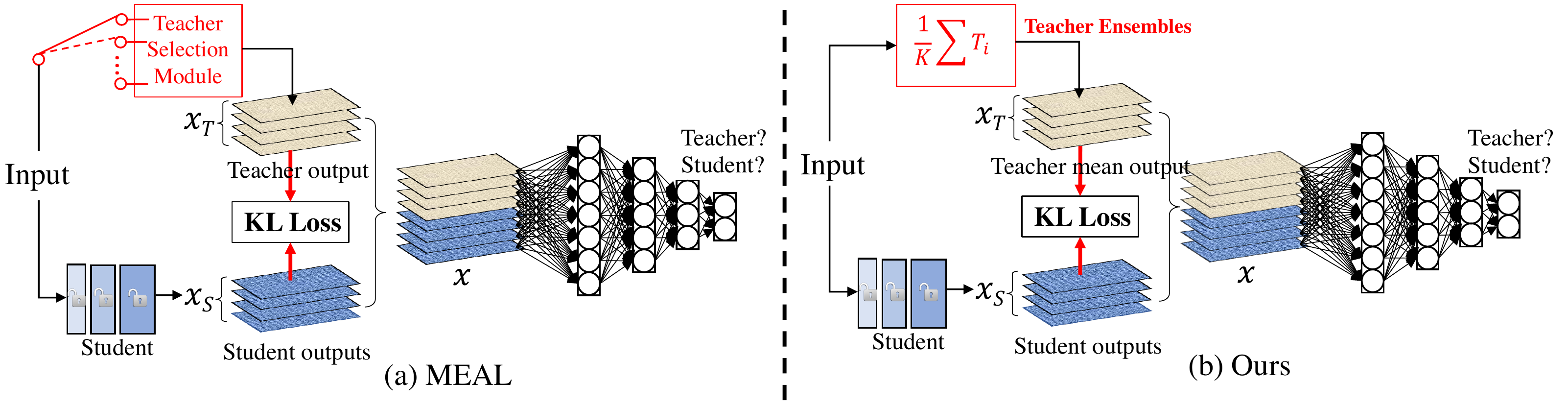}
  \vspace{-0.05in}
  \caption{An illustration of the comparison between MEAL~\cite{shen2019meal} and our method. We use an ensemble of all teacher networks instead of the teacher selection module as adopted in MEAL.}
  \label{fig:decay}
  \vspace{-0.1in}
\end{figure*}

\begin{table*}[t]
	\centering
	\caption{Item-by-item comparison of techniques that we use and do not use in our distillation training.}
	\resizebox{0.71\textwidth}{!}{
		\label{alternate}
		\begin{tabular}{l|c|c|c}
			\hline
			\multicolumn{2}{c}{\bf What we do not use}   & \multicolumn{2}{c}{\bf What we use} \\ \hline 
			architecture modification & \textcolor{red2}{\ding{56}}  &  an ensemble of giant pre-trained teachers & \textcolor{darkergreen}{\ding{52}} \\   \hline
			outside training data  & \textcolor{red2}{\ding{56}}  & KL divergence loss & \textcolor{darkergreen}{\ding{52}} \\   \hline
			hard/one-hot labels during distillation & \textcolor{red2}{\ding{56}}  & a good initialization for the student & \textcolor{darkergreen}{\ding{52}} \\   \hline
			cosine/linear decay learning rate & \textcolor{red2}{\ding{56}}  &  step decay/milestone learning rate (0.01-0.001) & \textcolor{darkergreen}{\ding{52}} \\   \hline
			weight-decay & \textcolor{red2}{\ding{56}}  &  &  \\   \hline
            cutout~\cite{devries2017improved}/mixup~\cite{zhang2017mixup}/cutmix~\cite{yun2019cutmix} training & \textcolor{red2}{\ding{56}}  &        &  \\   \hline
            label smoothing~\cite{szegedy2016rethinking} & \textcolor{red2}{\ding{56}}  &          &  \\   \hline            
            autoaug~\cite{cubuk2018autoaugment}/randaug~\cite{cubuk2020randaugment}, etc. & \textcolor{red2}{\ding{56}}  &         &  \\   \hline
            warmup~\cite{goyal2017accurate} & \textcolor{red2}{\ding{56}}  &         &  \\   \hline
		\end{tabular}
	}
	\vspace{-0.13in}
\end{table*}

\section{Framework Components}

In this section, we begin by introducing each component in our proposed framework, including: 1) teacher ensemble; 2) KL-divergence loss; 3) the discriminator. Then, we present the training details and techniques that we used and did not use in our distillation training.

\noindent{\textbf{Teachers Ensemble}} is used to generate more accurate predictions for guiding the student training. Different from MEAL~\cite{shen2019meal} that selected one teacher through a teacher selection module in each training iteration, we adopt the average of softmax probabilities from multiple pre-trained teachers as an ensemble. Let ${\mathcal{T}_\theta }$ be the teacher network, the output ensemble probability $\bf \hat p_e^{{\mathcal{T}_\theta }}$ can be described as:
\begin{equation}
    \begin{gathered}
    {\bf \hat p}_e^{{\mathcal{T}_\theta }}({X}) = \frac{1}{ K} \sum\limits_{{\bf t} = 1}^{ K} {\bf p_t}^{{\mathcal{T}_\theta }}({X})
    \end{gathered}
\end{equation}
where $\bf  p_{\bf t}^{{\mathcal{T}_\theta }}$ is the $\bf t$-th teacher's softmax prediction. $X$ is the inout image and $ K$ is the number of total teachers. $e$ denotes the ensembled probability.

\noindent{\textbf{KL-divergence}} is a measure metric of how one probability distribution is different from another reference distribution. In our approach, we train the student network $\mathcal{S}_\theta$ by minimizing the KL-divergence between its output ${\bf p}^{{\mathcal S}_\theta}({x_i})$ and the ensembled soft labels ${\bf \hat p}^{{\mathcal T}_\theta}({x_i})$ generated by the teacher ensemble. The loss function of KL-divergence can be formulated as (temperature is used as 1 following~\cite{shen2019meal}):
\begin{equation}
	\begin{gathered}
		{\LL_{KL}}({\mathcal{S}_\theta }) =  - \frac{1}{N} {\sum\limits_{i = 1}^N {{\bf \hat p}_e^{{\mathcal{T}_\theta }}({x_i})\log (\frac{{{\bf p}^{{S_\theta }}({x_i})}}{{{\bf \hat p}_e^{{\mathcal{T}_\theta }}({x_i})}})} }  \hfill 
	\end{gathered} 
\end{equation}
where $N$ is the number of samples. 
In practice, we can simply minimize the equivalent cross-entropy loss as follows:
\begin{equation}
	{{{\LL}}_{CE}}({\mathcal{S}_\theta }) =  - \frac{1}{N}{\sum\limits_{i = 1}^N {{\bf \hat p}_e^{{\mathcal{T}_\theta }}({x_i})\log } } {\bf p}^{{\mathcal{S}_\theta }}({x_i})
\end{equation}

\noindent{\textbf{Discriminator}} is a binary classifier to distinguish the input features are from teacher ensemble or student network. It consists of a {\em sigmoid} function following the {\em binary cross-entropy} loss. The loss can be formulated as: 
\begin{equation}
    \LL_{\mathcal D} =  - \frac{1}{N}\sum\limits_{i = 1}^N {\left[ {\mathbf{y}{_i} \cdot \log {\mathbf{\hat p^{\mathcal{D}}}_i}
    + (1 - \mathbf {y}{_i}) \cdot \log (1 - {\mathbf{\hat p^{\mathcal{D}}}_i})} \right]}
\end{equation}
where $\mathbf{y}_i$ is the binary label for the input features ${x}_i$, $\mathbf y \in \{0,1\}$, and $\mathbf{\hat p^{\mathcal{D}}}_i$ is the corresponding probability vector.

We define a {\em sigmoid} function to model the individual teacher or student probability:
\begin{equation}
\begin{gathered}
 \mathbf{\hat p^{\mathcal{D}}}({x};\theta) = \sigma (f_{\theta}(\{x_{\mathcal{T}},x_{\mathcal{S}}\}))
 \end{gathered} 
 \label{probability}
\end{equation}
where $f_{\theta}$ is a three-fc-layer subnetwork and $\theta$ is its parameters, $\sigma(x) = 1/( 1 + \exp(-x))$ is the logistic function.
In our model, we use the last output layer before softmax as the representation for the discriminator input.

Consider that our teacher supervision is an ensemble of multiple networks, it is not convenient to obtain the intermediate outputs. Also, to make the whole framework neater, we only adopt the similarity loss and discriminator on the final outputs of networks for distillation. We show from our experimental results that supervision from the last layer of teacher ensemble is competent to distill a strong student. 

\subsection{Model Capacity and Weight Decay}

Weight decay is a widely used regularization technique in neural networks while it is not used in our framework. It is worthwhile to discuss the motivation behind this choice. Weight decay is basically delivering the same effects as $L_2$ regularization. Since $L_2$ will penalize the large parameters in a network (as shown in Fig.~\ref{fig:value}), in our perspective, such an operation will impair the capacity of a network. We illustrate a comparison in Fig.~\ref{fig:decay} (right) of using weight decay and without it. It is curious to ask why previous models need weigh decay and it also seems helpful. We conjecture most of the previous networks are not yet saturated even those are trained with massive data augmentation and more training epochs, hence weight decay can help to prevent from overfitting and learn more information, the lost capacity is negligible and not so necessary. But for our initialized model, the performance is already high and we guess it is somewhat close to the upper bound of the network itself's capability, so the loss of capacity will be crucial for a network and weight decay may be harmful. Moreover, since our supervision from a strong teacher ensemble is fairly precise, the student should have enough capability to mimic such distribution, weight decay may not be applicable as it will reduce the complexity of a network. Also, the soft label itself in distillation has the regularization effect, so weight decay is not so necessary for distillation framework.

\section{Experiments}

\begin{table*}[h]
	\centering
	\caption{Comparison of validation accuracy on ImageNet dataset for ResNet-50 architecture under single crop evaluation. (*) indicates that they used horizontal flip, shifted center crop and color jittering for training.} 
	\resizebox{0.77\textwidth}{!}{
		\label{resnet50}
		\begin{tabular}{l|c|c|c|c}
			\hline
			\bf Network & Resolution & \#Params  & \bf Top-1 (\%)  & \bf Top-5  (\%) \\ \hline 
			ResNet-50 & 224 & 25.6M & 76.15 &92.86 \\   \hline
			ResNet-50 + DropBlock, (kp=0.9)~\cite{ghiasi2018dropblock}& 224 & 25.6M &78.13 & 94.02 \\  
            ResNet-50 + DropBlock (kp=0.9)~\cite{ghiasi2018dropblock} + label smoothing (0.1) & 224 & 25.6M  & 78.35  & 94.15 \\  
            ResNet-50 +  MEAL~\cite{shen2019meal} &  224 & 25.6M  & 78.21  &  94.01  \\ 
            \bf ResNet-50 + Ours (MEAL V2) & \cellcolor{mygraylite} 224 & 25.6M & \cellcolor{mygraylite} \bf 80.67  & \cellcolor{mygraylite} \bf 95.09  \\ \hline
            ResNet-50 + FixRes~\cite{touvron2019fixing}  & 384 & 25.6M & 79.0 &  94.6 \\
            ResNet-50 + FixRes (*)~\cite{touvron2019fixing}  & 384 & 25.6M & 79.1 & 94.6 \\
            \bf ResNet-50 + Ours (MEAL V2) & 380 & 25.6M & \bf 81.72  & \bf 95.81
            \\ \hline
            ResNet-50 + FixRes~\cite{touvron2019fixing} + CutMix  &  320 & 25.6M &  79.7 &  94.9 \\
            ResNet-50 + FixRes~\cite{touvron2019fixing} + CutMix (*) &  320 & 25.6M & 79.8 & 94.9 \\
            \bf ResNet-50 + Ours (MEAL V2) + CutMix & \cellcolor{mygray} \bf  224 & 25.6M  & \cellcolor{mygray} \bf 80.98  & \cellcolor{mygray} \bf 95.35
            \\ 
			\hline
		\end{tabular}
	}
	\vspace{-0.05in}
\end{table*}

\begin{table*}[h]
	\centering
	\caption{Comparison of validation accuracy on ImageNet  for MobileNet V3-Small 0.75/1.0/Large 1.0 and EfficientNet-B0 architectures.}
	\resizebox{0.58\textwidth}{!}{%
		\label{compact}
		\begin{tabular}{l|c|c|c|c}
			\hline
			\bf Network &  Resolution & \#Params  & \bf Top-1 (\%)  & \bf Top-5  (\%) \\ \hline 
			MobileNet V3-Small 0.75~\cite{howard2019searching}  & 224 & 2.04M & 65.40 & -- \\
            \bf + Ours (MEAL V2) & 224 & 2.04M & \bf 67.60 & \bf 87.23 \\ 
            MobileNet V3-Small 1.0~\cite{howard2019searching}  & 224 & 2.54M & 67.40 &  -- \\
            \bf + Ours (MEAL V2) & 224 & 2.54M & \bf 69.65 & \bf 88.71 \\ 
            MobileNet V3-Large 1.0~\cite{howard2019searching} & 224 &  5.48M & 75.20 &  -- \\ 
            \bf + Ours (MEAL V2) & 224  & 5.48M & \bf 76.92  & \bf  93.32 \\ \hline
            EfficientNet-B0 ~\cite{tan2019efficientnet}  &  224 & 5.29M & 77.3 (76.8) & 93.5 (93.2) \\ 
            \bf + Ours (MEAL V2) & 224 & 5.29M  & \bf 78.29 & \bf 93.95
            \\ 
			\hline
		\end{tabular}
	}
		\vspace{-0.1in}
\end{table*}

\noindent{\textbf{Main Dataset.}}
We conduct experiments on ILSVRC 2012 classification dataset~\cite{deng2009imagenet} that consists of 1,000 classes, with a number of 1.2 million training images and 50,000 validation images. We adopt the basic data augmentation scheme following~\cite{paszke2019pytorch}, i.e., {\em RandomResizedCrop} and {\em RandomHorizontalFlip}, and apply the single-crop operation at test time.

\noindent{\textbf{Transfer Learning Datasets.}} We study the transferability of our learned models on two mainstream tasks: the multiple-object/fine-grained classification and object detection. We conduct experiments on the following datasets: PASCAL VOC 2007~\cite{everingham2010pascal}, CUB200-2011~\cite{WahCUB_200_2011}, Birdsnap~\cite{berg2014birdsnap} and CIFAR-10~\cite{krizhevsky2009learning} for classification, and COCO~\cite{lin2014microsoft} with RetinaNet~\cite{lin2017focal} for detection.

\vspace{-0.05in}
\subsection{Experimental Settings}
\vspace{-0.05in}
We use a mini-batch size of 512 with 8 GPUs for training our models. SGD optimizer is adopted with a step learning rate decay scheduler. The initial learning rate is set to 0.01. We train with a total number of 180 epochs and the learning rate multiplied by 0.1 at 100 epoch. The weight decay is not used (set to 0) in our training. We apply this strategy to all our experiments regardless of what kind of teacher and student architectures we choose. We use the models in timm\footnote{\url{https://github.com/rwightman/pytorch-image-models/}.}. If the input size of a student network is $224\times224$, we choose senet154 and resnet152\_v1s as teachers according to the input size of the pre-trained models. For $380\times380$, we use efficientnet\_b4\_ns and efficientnet\_b4 as teachers. 

\begin{figure*}[h]
  \centering
  \includegraphics[width=0.99\textwidth]{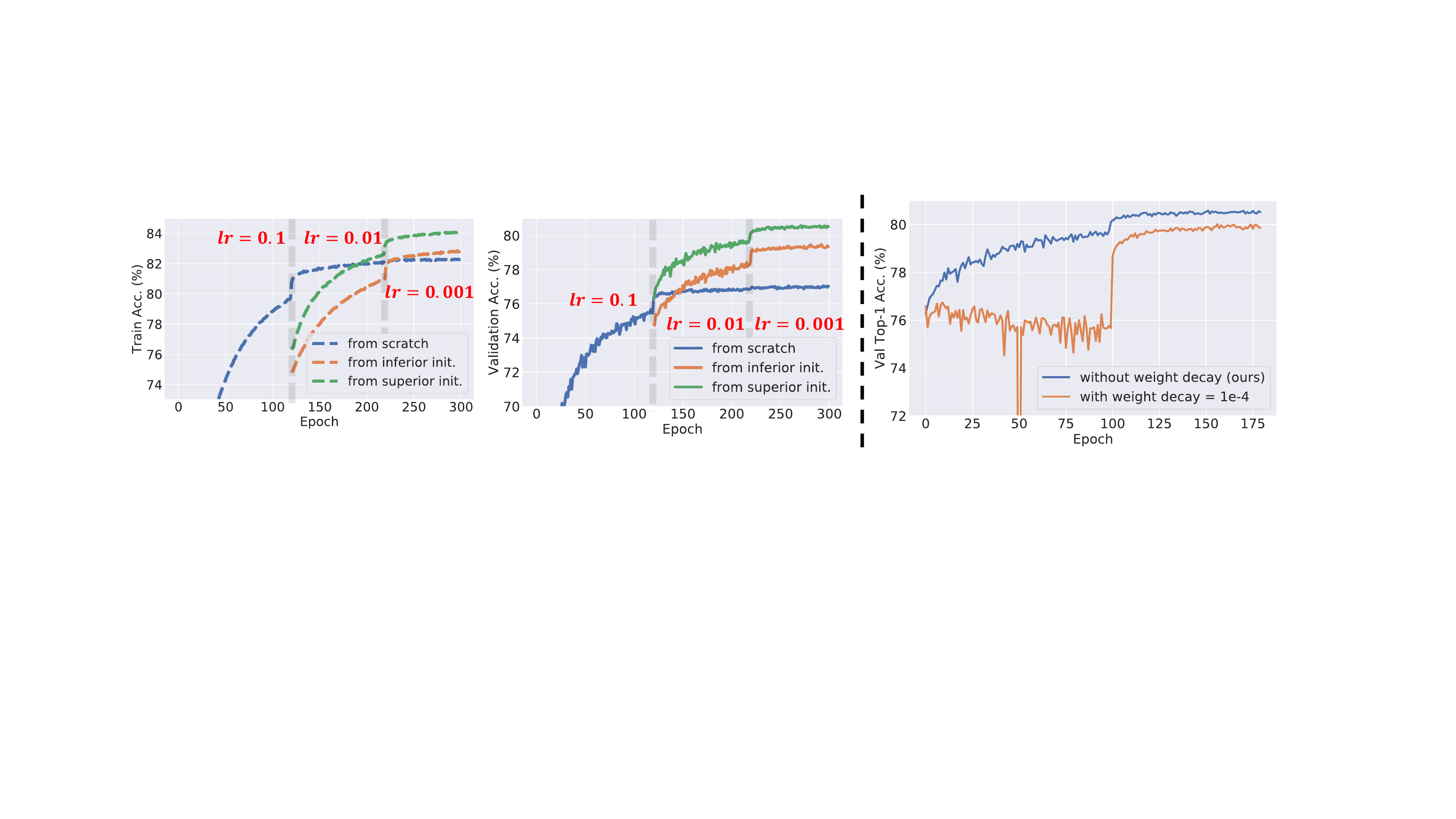}
  \vspace{-0.1in}
  \caption{A comparison of training from random initialized parameters, inferior parameters and superior parameters. Left are the training accuracy curves and middle are validation accuracy curves. The results indicate that a good initialization is crucial for the final performance. Left is the comparison of using weight decay (1e-4) and without it. We can observe that with weight decay, the training is surprisingly unstable in the first stage.}
  \label{fig:decay}
  \vspace{-0.15in}
\end{figure*}

\subsection{Main Results}

\noindent{\textbf{On ResNet-50.}} Our results on ResNet-50 are shown in Table~\ref{resnet50}. Under $224\times224$ input size, our method achieves 80.67\% Top-1 accuracy, outperforming the previous state-of-the-art method MEAL~\cite{shen2019meal} by 2.46\%. Furthermore, our results are even better than ResNeSt-50~\cite{zhang2020resnest} (fast) that requires to modify the network architecture and learned with many training tricks. After enlarging the input size to 380$\times$380, our performance is further improved to 81.72\%, outperforming FixRes (*)~\cite{touvron2019fixing} by 2.62\% with slightly smaller input.

\noindent{\textbf{On Small Networks.}} We choose MobileNet V3 Small-0.75/1.0/Large-1.0 and EfficientNet-B0 networks which are already compact models to verify the effectiveness of our proposed method. Our results are shown in Table~\ref{compact}, on MobileNet V3-Small 0.75 and 1.0, our method improves the original models by 2.20\% and 2.25\% accuracy without any architecture modification. Such huge increases are fairly surprising since the models are already compact, more importantly, the gains are totally free during inference stage. On MobileNet V3-Large 1.0 and EfficientNet-B0, although the improvement is not as large as Small 0.75 and 1.0, we still obtain 1.72\% and 1.49\% increase on ImageNet. Note that for EfficientNet-B0, 77.3/93.5 accuracy is from their paper~\cite{tan2019efficientnet} and 76.8/93.2 is the actual accuracy from their pre-trained models in timm.

\noindent{\textbf{With More Data Augmentation.}} We'd like to further explore whether our models have been saturated on the target data by injecting more data augmentation like CutMix in training. The results are shown in Table~\ref{resnet50}, we involve CutMix and keep other settings the same as our basic experiments, we obtain Top-1/5 80.98\%/95.35\%, which outperform the baseline MEAL V2 by 0.31\%/0.26\%. While the improvement is not so large, it indicates that our model is not yet over-fitting and still has room to boost. Moreover, our results are 1.18\%/0.45\% better than FixRes+CutMix (*) under smaller input resolution (224 vs. 320). Intriguingly, the results on ResNet-50 are very close to the teachers we used in distillation (81.38\%/95.39\% and 80.86\%/95.35\%), since the scale of our student is much smaller than the teacher architectures, it's surprising that the student can catch up the teachers without additional training data.

\noindent{\textbf{Criterion for Choosing Teachers.}} One critical factor for choosing teacher networks is the accuracy. From our experiments, it shows that stronger teachers usually distill better students. Another factor is the training settings on teachers, such as input resolution, image color space, etc. These setting should match the ones used during distillation so that the teachers can provide correct probability of the input as the supervision for students. Data augmentation can be different between the stages of training teachers and distillation, like we can use CutMix to train teachers but not use it in distillation, vice versa.

\begin{table}[h]
\vspace{-0.07in}
\caption{Comparison with other state-of-the-art knowledge distillation methods. We show the gap between teacher and student to demonstrate the mimicking ability of each method. T$_E$ denotes the ensembled result of our teachers. $^{\dagger}$ denotes training with CutMix.}
\label{tab:my-table_kd}
\resizebox{0.48\textwidth}{!}{%
\begin{tabular}{c|c|c|c}
\hline
\bf Method                 & \bf Teacher(s)                                                            & \bf {Student}              & \bf Gap with Teacher      \\ \hline
CRD (ICLR'20)~\cite{tian2019contrastive}   & 73.31/91.42         & {71.17/90.13}          &      $\Delta$2.14/1.39       \\ 
CRD+KD (ICLR'20)~\cite{tian2019contrastive}   & 73.31/91.42         & {71.38/90.49}          &      $\Delta$1.93/0.93       \\ \hline
AE-KD (NeurIPS'20)~\cite{du2020agree} (\#1)        &            75.67/92.50                      & 67.81/88.21        &  $\Delta$7.86/4.29             \\ 
AE-KD (NeurIPS'20)~\cite{du2020agree} (\#3)        &           76.85/93.60                       & 68.28/88.21         &  $\Delta$8.57/5.39           \\ 
AE-KD (NeurIPS'20)~\cite{du2020agree} (\#5)        &           77.52/93.85                       & 69.14/88.93        &  $\Delta$8.38/4.92           \\ \hline
\bf Ours (MEAL V2)                  & \begin{tabular}[c]{@{}c@{}}T$1$: 81.38/95.39\\T$2$: 80.86/95.35\\T$_E$: 82.67/96.13\end{tabular} & {{80.67/95.09}} &   \begin{tabular}[c]{@{}c@{}}$\Delta$0.71/0.30 \\  $\Delta$0.19/0.26 \\ \bf $\Delta$2.00/1.04 \end{tabular}     \\ \hline
\bf Ours (MEAL V2)$^{\dagger}$                   & \begin{tabular}[c]{@{}c@{}}T1: 81.38/95.39\\ T2: 80.86/95.35\\T$_E$: 82.67/96.13\end{tabular} & {{80.98/95.35}} &   \begin{tabular}[c]{@{}c@{}}$\Delta$0.40/0.04\\ $\Delta$-0.13/0.00  \\ \bf $\Delta$1.69/0.78 \end{tabular}    \\ \hline
\end{tabular}
\vspace{-0.16in}
}
\end{table}
\noindent{\textbf{Comparison with Other Distillation Methods.}}
We compare our distillation results with other state-of-the-art distillation methods in Table~\ref{tab:my-table_kd} on two aspects: (1) the mimicking ability through comparing the performance gap between teacher and student. Such comparison can directly reflect the true superiority and learning ability, no matter what kind of structures on teacher and student you use. It can be observed that our gaps between teachers and students are smaller than both CRD~\cite{tian2019contrastive} and AE-KD~\cite{du2020agree} by a significant margin. (2) the absolute accuracy on the same structure of student. If choosing ResNet-18 as the student, our method obtains 73.19\% Top-1 accuracy, outperforming CRD~\cite{tian2019contrastive} by 2.02\%. Although different methods used different architectures as teachers, like AE-KD~\cite{du2020agree} and our method applied multiple network ensemble as a teacher, such result still can verify the effectiveness of our method in a certain extent.

\vspace{-0.05in}
\subsection{Ablation Study and Analysis}
\vspace{-0.03in}

There are many factors in knowledge distillation that determines the performance of a student. Since we use the same teacher ensemble for all ResNet-50, MobileNet V3 and EfficientNet-B0 under 224$\times$224 input, the results indicate that the student architecture or capacity itself is a crucial factor. If we compare MEAL V1 and V2 we can further derive the conclusion that teacher's performance, i.e. the quality of supervision, is another factor for the student, generally, the stronger teachers can consistently distill stronger students. In the following, we would like to exam the impact of the initialization and discriminator.

\begin{figure*}[t]
  \centering
  \includegraphics[width=0.85\textwidth]{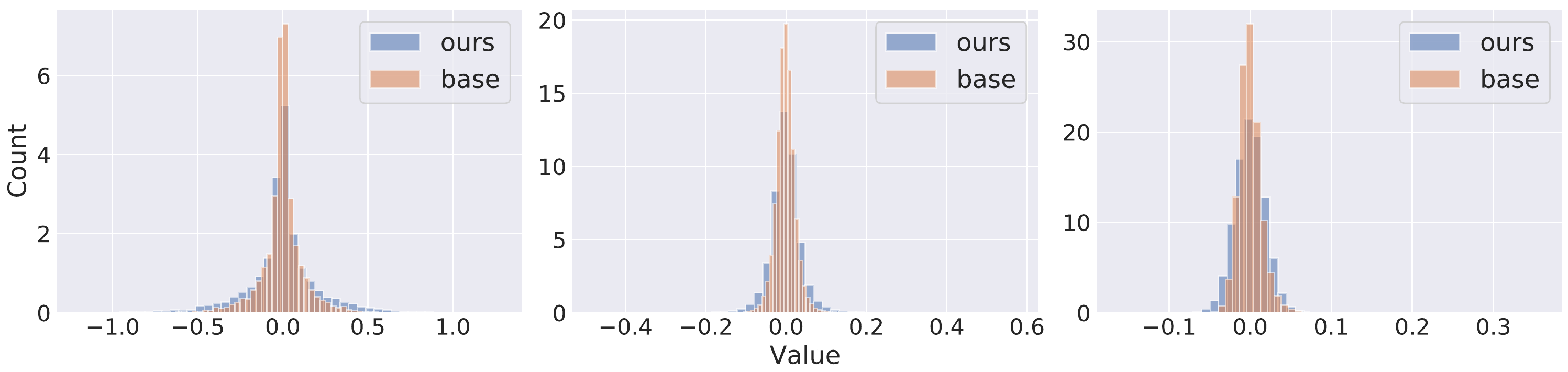}
  \vspace{-0.08in}
  \caption{An illustration of weight distributions in the first, middle ({last conv in block 2}) and last conv layers in a ResNet-50 model.} 
  \label{fig:distribution}
  \vspace{-0.15in}
\end{figure*}

\noindent{\textbf{Effects of initialization.}}
To verify whether the initialization of a student has a big impact, we conduct the ablation study through adopting (1) randomly initialized weights, (2) pytorch moderate weights and (3) timm superior weights. Results are illustrated in Fig.~\ref{fig:decay}, we trained with additional 120 epochs with $lr=0.1$ if the parameters are randomly initialized. Intriguingly, the convergence with randomly initialized weights is not as good as using pre-trained parameters, especially the second and third stages when $lr$ is smaller. The final accuracy (77.07\%) is only slightly better than that with hard label. Adopting the standard one-hot label pre-trained weights can dramatically improve the results to 79.48\%, but is still slightly worse than the superior initialization. Models in timm are trained with massive data augmentation techniques so the accuracy is higher. It seems that our framework can inherit the knowledge in such initialization and promotes it to a better status of student. Since the good initialization essentially is learned from hard label, the knowledge learned by hard label is complementary with that learned from soft label. When training from scratch, the hard label is suggested to be involved with a pre-training stage to encode more information even it is not accurate or strong, then remove it in the latter part of distillation and inherit weights with soft-label solely for better guidance.

We also examine the sensitivity of initialization for the final performance. We chose {\em tf\_efficientnet\_b0} (Top-1/5: 76.85\%/93.25\%) and {\em efficientnet\_b0} (77.70\%/93.53\%) as the student initialization in timm, respectively. They have the same architecture but the performance is different due to the discrepant training settings. Interestingly, we got Top-1 78.29\% and 78.23\% respectively for the two initializations with the same teacher ensembles and training hyper-parameters. It indicates that the final performance is not sensitive to the subtle difference of good initializations.

\noindent{\textbf{With or w/o the discriminator.}}
The discriminator is used to prevent the student from being overfitting on the training data. It can slow down the moving of a student to mimic the teachers' output, which can be regarded as a regularization effect. 
In the scenario of MEAL V2, our teacher ensembles are usually powerful and strong, meanwhile, the student architectures are usually smaller and more compact than the teachers, meaning that the capability and learning ability are also much worse than the pretrained teacher networks, even we force the student to produce the same predictions as strong teachers, the outputs between student and teacher ensembles still have inevitable gaps which cannot be eradicated through the KL-divergence loss. That is to say, the discriminator is very easy to distinguish that the feature is from a student or teacher ensemble and the regularization effect will be weakened. The performance comparison of using and without the discriminator is shown in our Appendix. 
Even so, in MEAL V2 we still see slight improvement on performance by using the discriminator.

\begin{figure}[t]
  \centering
  \vspace{0.01in}
  \includegraphics[width=0.50\textwidth]{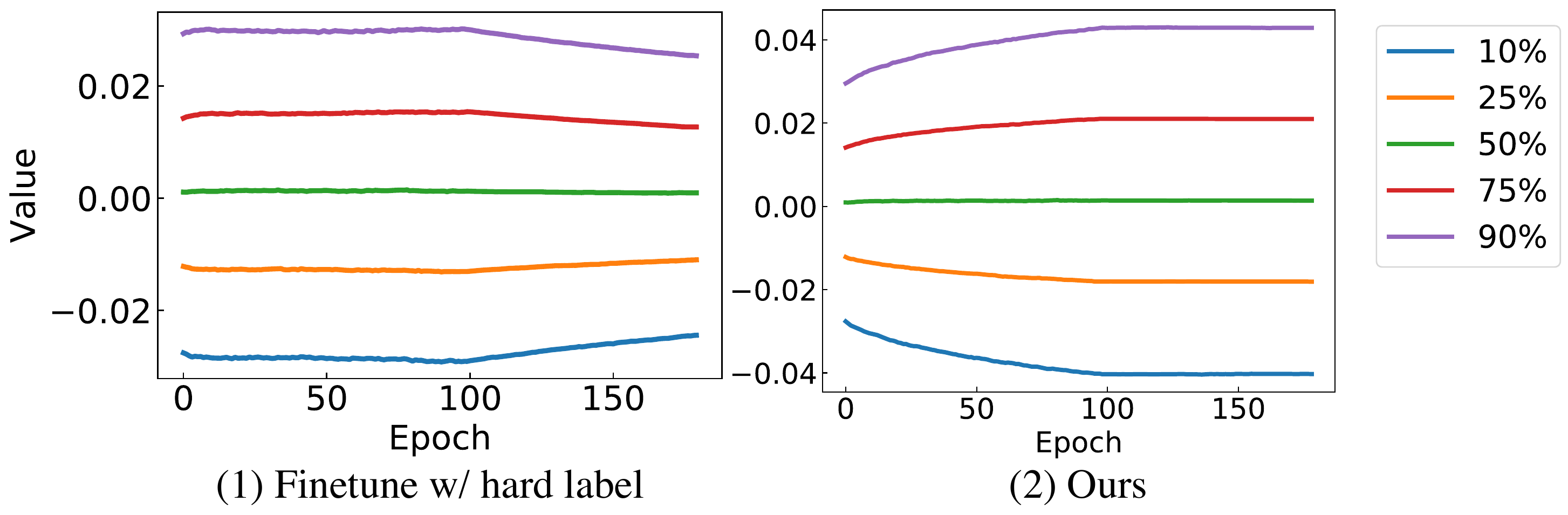}
  \vspace{-0.19in}
  \caption{Evolution of weights percentile. We monitor the middle convolutional layer ({last conv in block 2}) in ResNet-50 with elements at 10\%, 25\%, 50\%, 75\%, 90\% during whole training.}
  \label{fig:value}
  \vspace{-0.16in}
\end{figure}

\vspace{-0.05in}
\subsection{Visualization}
\vspace{-0.03in}

We provide two visualizations schemes to understand why our trained model can achieve substantially better results. The first is the histogram of weights in particular convolutional layers as shown in Fig.~\ref{fig:distribution}, we choose the pytorch pre-trained model for a comparison. We can see our weights always have a wider scope of values and fewer elements are close to zero. We argue the behavior of wider scope on weights reflects the larger capacity of a network since weights can have more potential values or status to be. Our second is the evolution of weights percentile during training, as in Fig.~\ref{fig:value}. The reference is a model trained with one-hot labels and both of them use the same initialization. Consistently, we observe the values of weights tend to diffuse rather than the convergence. We conjecture this is caused by weight decay as we have discussed earlier.

\vspace{-0.05in}
\section{Transfer Learning}
\vspace{-0.03in}

\noindent{\textbf{Fine-tuning backbone.}} On the classification task, we fine-tune the entire network of ResNet-50 using the parameters of the pretrained model as initialization. Since PASCAL VOC classification is a multi-label problem, we apply sigmoid cross-entropy objective for it, and softmax cross-entropy for other datasets. On COCO detection with RetinaNet~\cite{lin2017focal}, we use exactly the same hyper-parameters in detectron2~\cite{wu2019detectron2} but replacing the initialization with ours.

\begin{table}[t]
\centering
\vspace{-0.05in}
\caption{Transfer accuracy on classification task.}
\label{tab:my-transfer-cls}
	\resizebox{0.42\textwidth}{!}{
\begin{tabular}{lcccc}
 \hline
 & VOC2007 & CUB200-2011 & Birdsnap & CIFAR-10  \\
            \hline
\em From Scratch & 72.20& 58.90    &  66.97 &   94.71     \\ \hline
\multicolumn{5}{l}{\em Fine-tune:}                              \\
Base model  & 94.00 & 80.70  &  75.28  &  97.24    \\
Ours        & \bf 95.10 &\bf 83.70  &  \bf 75.55  &  \bf 97.26   \\ \hline
\multicolumn{5}{l}{\em Freeze backbone:}                            \\
Base model  & 87.50 &  66.27    &    \bf 55.01   &  81.69   \\
Ours        & \bf 90.80 & \bf 68.29   &   53.56   & \bf 84.16 \\\hline
\end{tabular}
}
\vspace{-0.15in}
\end{table}

\begin{table}[t]
\caption{Transfer accuracy on COCO detection using RetinaNet.}
\label{tab:my-transfer-det}
	\resizebox{0.45\textwidth}{!}{
\begin{tabular}{lcccccc}
 \hline
 & AP & AP$_{50}$ & AP$_{75}$ & AP$_{Small}$ & AP$_{Medium}$ & AP$_{Large}$ \\
            \hline
\multicolumn{5}{l}{\em Fine-tune all layers:}                              \\
Base model  &   37.234    & 56.436 & 39.769 & \bf 22.899  &  41.167&47.576    \\
Ours        &  \bf 37.501  & \bf 56.829  & \bf 40.148  & 21.483 & \bf 41.264 & \bf 48.655 \\ \hline
\multicolumn{5}{l}{\em Freeze first stage of backbone:}                            \\
Base model  &37.253 & 56.636 & 39.912 & 22.143 &\bf 41.567 & 47.322 \\
Ours        &\bf 37.501 & \bf 56.850 & \bf 40.245 & \bf 22.471 & 41.410 &\bf 48.853  \\\hline
\end{tabular}
}
\vspace{-0.2in}
\end{table}

\noindent{\textbf{Fixing backbone.}} We freeze the entire backbone and solely train the last linear layer. This is the linear evaluation to verify the quality of learned representations. 
For detection task, we freeze the first stage of backbone instead of the entire network. More details on finetuning and linear evaluation will be given in Appendix. 
Our results are shown in Table~\ref{tab:my-transfer-cls} and \ref{tab:my-transfer-det}. In the most cases and datasets, a consistent improvement is achieved by using our trained parameters.

\vspace{-0.05in}
\section{Conclusion}
\vspace{-0.03in}

We have presented a new paradigm of knowledge distillation based on a teacher ensemble and a discriminator. We show that such a simple framework can achieve promising results  without tricks on a variety of network structures including the extremely tiny and compact models. On ImageNet, our method achieves {\bf 80.67\%} top-1 accuracy using a single crop of 224$\times$224 on the vanilla ResNet-50. Our results show that existing networks' potential has not been fully exploited and there is still room to boost and enhance through our framework. We hope the proposed method can inspire more studies along this direction of boosting tiny and compact models through knowledge distillation.

\appendix
\newpage

{\Large \noindent{\textbf{Appendix}}}

\section{Implementation Details of Transferring}

\noindent{\textbf{Fine-tuning backbone.}} On classification task, we fine-tune the entire network of ResNet-50 using the parameters of the pre-trained model as initialization. We train for 200 epochs with a batch size of 128 and an initial learning rate of 0.01. Since PASCAL VOC classification is a multi-label problem, we apply sigmoid cross-entropy objective for it, and softmax cross-entropy for other datasets. We use SGD with a momentum parameter of 0.9 and weight decay of 0.0001. We perform standard random crops with resize and flips as data augmentation during fine-tuning. The training image size is 224$\times$224, at test time, we resize images to 256 pixels and take a 224$\times$224 center crop. On COCO detection with RetinaNet~\cite{lin2017focal}, we use exactly the same hyper-parameters in detectron2~\cite{wu2019detectron2} but replacing the initialization with ours. 

\noindent{\textbf{Fixing backbone.}} We freeze the entire backbone and solely train the last linear layer. This is the linear evaluation protocol to verify the quality of learned representations. We adopted the same training setting as fine-tuning but using a larger initial learning rate of 0.1. For detection task, we freeze the first stage of backbone instead of the entire network and follow the experimental settings of detectron2~\cite{wu2019detectron2}.

\noindent{\textbf{Overview of transfer learning datasets.}} As shown in Table~\ref{tab:my-table_dataset}, we provide a brief overview of five datasets that we used in our transfer learning experiments.

\begin{table}[h]
\centering
\caption{Overview of five datasets used in our experiments for transfer learning.}
\label{tab:my-table_dataset}
\resizebox{0.5\textwidth}{!}{
\begin{tabular}{l|c|c|c|c}
\toprule[1.5pt]
\bf Dataset & \#class & Property          & \#Train(+val) set & \#Testing set \\ \midrule \midrule
VOC 2007~\cite{everingham2010pascal}    &  20    & Multi-object     & 5,011 & 4,952 \\  
CUB200-2011~\cite{WahCUB_200_2011}    &  200     & Fine-grained  & 5,994 & 5,794 \\
Birdsnap~\cite{berg2014birdsnap}   &  500     & Fine-grained   & 47,386   &  2,443\\ 
CIFAR-10~\cite{krizhevsky2009learning} & 10  & Standard    & 50,000 & 10,000  \\
MS COCO~\cite{lin2014microsoft}    &  80    & Detection  & 123,287 & 40,775 \\
\bottomrule[1.5pt]
\end{tabular}
}
\end{table}

\section{More Comparison}

\noindent{\textbf{With or w/o the discriminator.}}
As we have introduced in the main text, the discriminator is used to prevent the student from being overfitting on the training data. It can slow down the moving of a student to mimic the teachers' output, which can be regarded as a regularization effect. 
In the scenario of MEAL V2, our teacher ensembles are usually large and heavy, while, the student architectures are smaller and more compact than the teachers, meaning that the capability and learning ability are also much worse than the pretrained teacher networks, even we force the student to produce the same predictions as strong teachers, the outputs between student and teacher ensembles still have inevitable gaps which cannot be eradicated through the KL-divergence loss. That is to say, the discriminator is easy to distinguish that the feature is from a student or teacher ensemble and the regularization effect from the discriminator will be weakened. 
Fig.~\ref{fig:dis_com} illustrates the performance comparison with and without the discriminator. Even so, in MEAL V2 we still see slight improvement by using the discriminator.

\begin{figure}[t]
  \centering
  \includegraphics[width=0.4\textwidth]{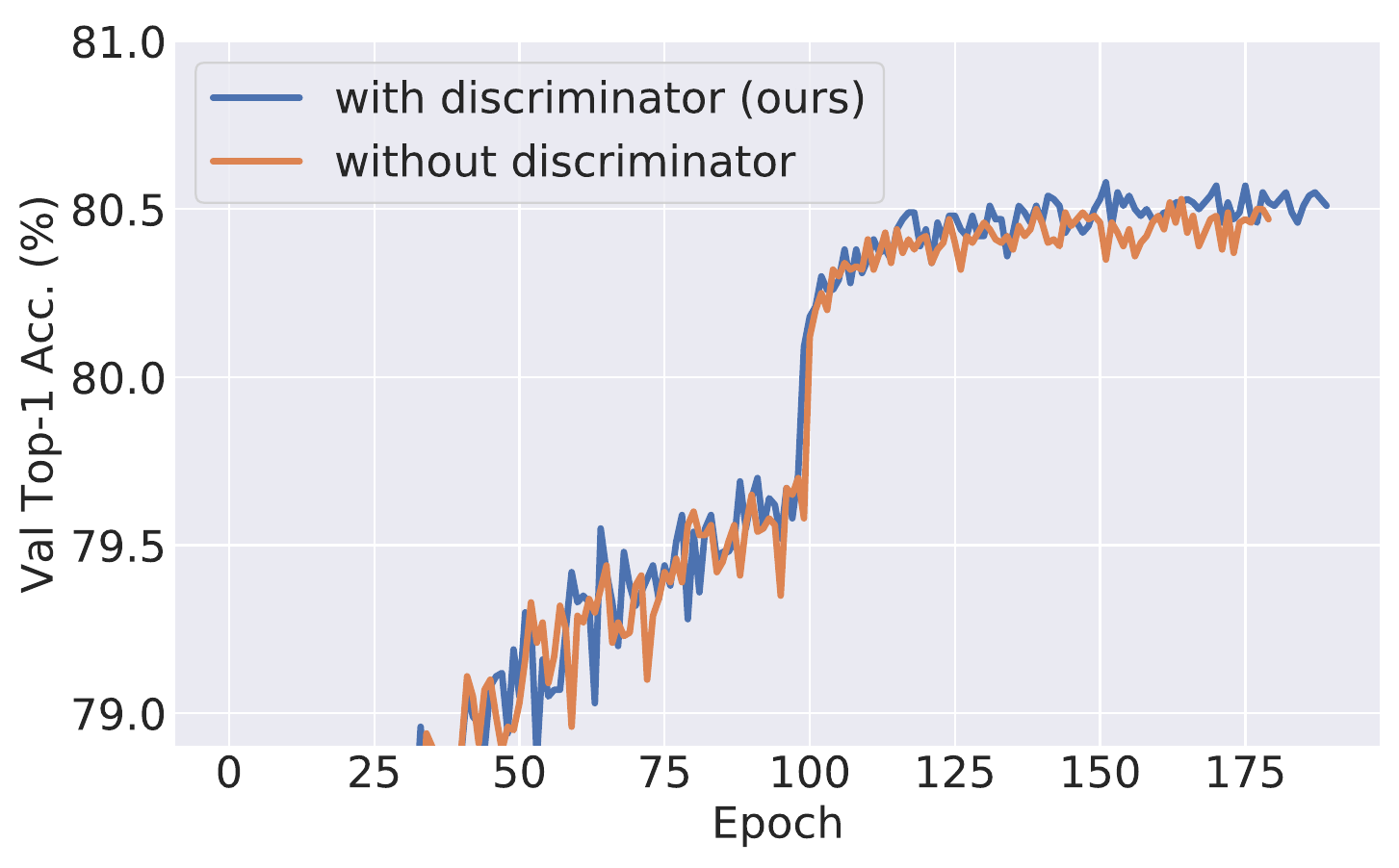}
  \vspace{-0.08in}
  \caption{Comparison of Top-1 accuracy with and without the discriminator on ImageNet validation set.} 
  \label{fig:dis_com}
  \vspace{-0.15in}
\end{figure}

\section{Discussions} \label{discussion}

\noindent{\textbf{Why is the hard/one-hot label not necessary in knowledge distillation?}}
The one-hot labels in ImageNet are annotated by humans, thus there are certainly some incorrect or missing annotations into them. Also, a non-negligible proportion of images in ImageNet contain more than one object within a single image, the one-hot label is determined by the annotators among multiple objects which cannot represent the complete content of this image precisely. We argue that if the teacher ensembles are strong enough, which can provide high-quality predictions for the input image, involving such inaccurate hard labels will mislead the student to a wrong optimum and incur inferior performance. Moreover, the distilled soft labels can overcome the noise and mismatching issues caused by the random crop data augmentation strategy adopted in deep model training.

\noindent{\textbf{Training w/ and w/o the good initialization.}} As we mentioned in the main paper, training without the good initialization obtained inferior performance. However, after involving the hard labels and training the initialization with the hard labels by standard settings, our distillation framework can boost such initialization model by $\sim$3\% and the final performance is competitive. This procedure indicates that hard labels and soft labels are complementary. Also, it is equivalent to our proposed framework since our good initialization is trained with hard labels. Hence, our framework can be regarded as a new procedure: {\em hard label pre-training + soft label finetuning}.

\noindent{\textbf{How about the generalization ability of our method on large students?}} We tried to use some large models like ResNeXt-101 32$\times$48d for the students as used in teacher networks, which means that the student has similar capability with teachers. As expected, the improvement is not as considerable as those of small students, we still see some increase on performance. Generally, the soft supervision from teacher ensembles is better than the human-annotated hard labels. Especially when the scale and performance gap between teachers and students are enormous, the improvement will be more effective and notable. That is to say, in most of our experimental cases, the stronger teachers can consistently produce and distill stronger students.

\noindent{\textbf{Is there still room to improve the performance of vanilla ResNet-50?}} It's definitely {\em Yes}. Replacing the teacher ensembles we used with more and stronger networks could be helpful, but the training cost will be increased accordingly. Also, some of the common tricks like cosine decay learning rate might be useful for the performance but it needs more resources to verify and the framework will become not neat. The current choices are the compromise and a trade-off under the considerations of training efficiency, computational resources, etc. Our purpose of this paper is primarily to verify the effectiveness of our proposed perspective, rather than the highest accuracy. Still, it will be very interesting to explore the upper bound performance of a fixed-structure network, such as ResNet-50.

\noindent{\textbf{The relationship to the lottery ticket hypothesis~\cite{frankle2018lottery}.} \label{lottery} Lottery ticket hypothesis assumes that we can find a sub-network from a trained giant model, and retraining such initialization of subnetwork can scale its accuracy back to the original giant model. We also observe that the initialization of the student is crucial in our framework for the super teachers to play a role in knowledge distillation process. While, the difference from lottery ticket hypothesis is that our student is trained solely, rather than being selected as a sub-network from a large pre-trained network.


{\small
\bibliographystyle{ieee_fullname}
\bibliography{egbib}

\begin{thebibliography}{10}\itemsep=-1pt

\bibitem{berg2014birdsnap}
Thomas Berg, Jiongxin Liu, Seung Woo~Lee, Michelle~L Alexander, David~W Jacobs,
  and Peter~N Belhumeur.
\newblock Birdsnap: Large-scale fine-grained visual categorization of birds.
\newblock In {\em Proceedings of the IEEE Conference on Computer Vision and
  Pattern Recognition}, pages 2011--2018, 2014.

\bibitem{courbariaux2016binarized}
Matthieu Courbariaux, Itay Hubara, Daniel Soudry, Ran El-Yaniv, and Yoshua
  Bengio.
\newblock Binarized neural networks: Training deep neural networks with weights
  and activations constrained to+ 1 or-1.
\newblock {\em arXiv preprint arXiv:1602.02830}, 2016.

\bibitem{cubuk2018autoaugment}
Ekin~D Cubuk, Barret Zoph, Dandelion Mane, Vijay Vasudevan, and Quoc~V Le.
\newblock Autoaugment: Learning augmentation policies from data.
\newblock {\em arXiv preprint arXiv:1805.09501}, 2018.

\bibitem{cubuk2020randaugment}
Ekin~D Cubuk, Barret Zoph, Jonathon Shlens, and Quoc~V Le.
\newblock Randaugment: Practical automated data augmentation with a reduced
  search space.
\newblock In {\em Proceedings of the IEEE/CVF Conference on Computer Vision and
  Pattern Recognition Workshops}, pages 702--703, 2020.

\bibitem{deng2009imagenet}
Jia Deng, Wei Dong, Richard Socher, Li-Jia Li, Kai Li, and Li Fei-Fei.
\newblock Imagenet: A large-scale hierarchical image database.
\newblock In {\em 2009 IEEE conference on computer vision and pattern
  recognition}, pages 248--255. Ieee, 2009.

\bibitem{devries2017improved}
Terrance DeVries and Graham~W Taylor.
\newblock Improved regularization of convolutional neural networks with cutout.
\newblock {\em arXiv preprint arXiv:1708.04552}, 2017.

\bibitem{du2020agree}
Shangchen Du, Shan You, Xiaojie Li, Jianlong Wu, Fei Wang, Chen Qian, and
  Changshui Zhang.
\newblock Agree to disagree: Adaptive ensemble knowledge distillation in
  gradient space.
\newblock {\em Advances in Neural Information Processing Systems}, 33, 2020.

\bibitem{everingham2010pascal}
Mark Everingham, Luc Van~Gool, Christopher~KI Williams, John Winn, and Andrew
  Zisserman.
\newblock The pascal visual object classes (voc) challenge.
\newblock {\em International journal of computer vision}, 88(2):303--338, 2010.

\bibitem{frankle2018lottery}
Jonathan Frankle and Michael Carbin.
\newblock The lottery ticket hypothesis: Finding sparse, trainable neural
  networks.
\newblock {\em arXiv preprint arXiv:1803.03635}, 2018.

\bibitem{ghiasi2018dropblock}
Golnaz Ghiasi, Tsung-Yi Lin, and Quoc~V Le.
\newblock Dropblock: A regularization method for convolutional networks.
\newblock In {\em Advances in Neural Information Processing Systems}, pages
  10727--10737, 2018.

\bibitem{girshick2014rich}
Ross Girshick, Jeff Donahue, Trevor Darrell, and Jitendra Malik.
\newblock Rich feature hierarchies for accurate object detection and semantic
  segmentation.
\newblock In {\em Proceedings of the IEEE conference on computer vision and
  pattern recognition}, pages 580--587, 2014.

\bibitem{goyal2017accurate}
Priya Goyal, Piotr Doll{\'a}r, Ross Girshick, Pieter Noordhuis, Lukasz
  Wesolowski, Aapo Kyrola, Andrew Tulloch, Yangqing Jia, and Kaiming He.
\newblock Accurate, large minibatch sgd: Training imagenet in 1 hour.
\newblock {\em arXiv preprint arXiv:1706.02677}, 2017.

\bibitem{han2015deep}
Song Han, Huizi Mao, and William~J Dally.
\newblock Deep compression: Compressing deep neural networks with pruning,
  trained quantization and huffman coding.
\newblock {\em arXiv preprint arXiv:1510.00149}, 2015.

\bibitem{han2015learning}
Song Han, Jeff Pool, John Tran, and William Dally.
\newblock Learning both weights and connections for efficient neural network.
\newblock In {\em Advances in neural information processing systems}, pages
  1135--1143, 2015.

\bibitem{he2016deep}
Kaiming He, Xiangyu Zhang, Shaoqing Ren, and Jian Sun.
\newblock Deep residual learning for image recognition.
\newblock In {\em Proceedings of the IEEE conference on computer vision and
  pattern recognition}, pages 770--778, 2016.

\bibitem{he2017channel}
Yihui He, Xiangyu Zhang, and Jian Sun.
\newblock Channel pruning for accelerating very deep neural networks.
\newblock In {\em Proceedings of the IEEE International Conference on Computer
  Vision}, pages 1389--1397, 2017.

\bibitem{hinton2015distilling}
Geoffrey Hinton, Oriol Vinyals, and Jeff Dean.
\newblock Distilling the knowledge in a neural network.
\newblock {\em arXiv preprint arXiv:1503.02531}, 2015.

\bibitem{howard2019searching}
Andrew Howard, Mark Sandler, Grace Chu, Liang-Chieh Chen, Bo Chen, Mingxing
  Tan, Weijun Wang, Yukun Zhu, Ruoming Pang, Vijay Vasudevan, et~al.
\newblock Searching for mobilenetv3.
\newblock In {\em Proceedings of the IEEE International Conference on Computer
  Vision}, pages 1314--1324, 2019.

\bibitem{howard2017mobilenets}
Andrew~G Howard, Menglong Zhu, Bo Chen, Dmitry Kalenichenko, Weijun Wang,
  Tobias Weyand, Marco Andreetto, and Hartwig Adam.
\newblock Mobilenets: Efficient convolutional neural networks for mobile vision
  applications.
\newblock {\em arXiv preprint arXiv:1704.04861}, 2017.

\bibitem{hu2018squeeze}
Jie Hu, Li Shen, and Gang Sun.
\newblock Squeeze-and-excitation networks.
\newblock In {\em Proceedings of the IEEE conference on computer vision and
  pattern recognition}, pages 7132--7141, 2018.

\bibitem{huang2017densely}
Gao Huang, Zhuang Liu, Laurens Van Der~Maaten, and Kilian~Q Weinberger.
\newblock Densely connected convolutional networks.
\newblock In {\em Proceedings of the IEEE conference on computer vision and
  pattern recognition}, pages 4700--4708, 2017.

\bibitem{hubara2017quantized}
Itay Hubara, Matthieu Courbariaux, Daniel Soudry, Ran El-Yaniv, and Yoshua
  Bengio.
\newblock Quantized neural networks: Training neural networks with low
  precision weights and activations.
\newblock {\em The Journal of Machine Learning Research}, 18(1):6869--6898,
  2017.

\bibitem{jacob2018quantization}
Benoit Jacob, Skirmantas Kligys, Bo Chen, Menglong Zhu, Matthew Tang, Andrew
  Howard, Hartwig Adam, and Dmitry Kalenichenko.
\newblock Quantization and training of neural networks for efficient
  integer-arithmetic-only inference.
\newblock In {\em Proceedings of the IEEE Conference on Computer Vision and
  Pattern Recognition}, pages 2704--2713, 2018.

\bibitem{jegou2017one}
Simon J{\'e}gou, Michal Drozdzal, David Vazquez, Adriana Romero, and Yoshua
  Bengio.
\newblock The one hundred layers tiramisu: Fully convolutional densenets for
  semantic segmentation.
\newblock In {\em Proceedings of the IEEE conference on computer vision and
  pattern recognition workshops}, pages 11--19, 2017.

\bibitem{krizhevsky2009learning}
Alex Krizhevsky et~al.
\newblock Learning multiple layers of features from tiny images.
\newblock 2009.

\bibitem{krizhevsky2012imagenet}
Alex Krizhevsky, Ilya Sutskever, and Geoffrey~E Hinton.
\newblock Imagenet classification with deep convolutional neural networks.
\newblock In {\em Advances in neural information processing systems}, pages
  1097--1105, 2012.

\bibitem{kuznetsova2018open}
Alina Kuznetsova, Hassan Rom, Neil Alldrin, Jasper Uijlings, Ivan Krasin, Jordi
  Pont-Tuset, Shahab Kamali, Stefan Popov, Matteo Malloci, Tom Duerig, et~al.
\newblock The open images dataset v4: Unified image classification, object
  detection, and visual relationship detection at scale.
\newblock {\em arXiv preprint arXiv:1811.00982}, 2018.

\bibitem{lecun1998gradient}
Yann LeCun, L{\'e}on Bottou, Yoshua Bengio, and Patrick Haffner.
\newblock Gradient-based learning applied to document recognition.
\newblock {\em Proceedings of the IEEE}, 86(11):2278--2324, 1998.

\bibitem{li2016pruning}
Hao Li, Asim Kadav, Igor Durdanovic, Hanan Samet, and Hans~Peter Graf.
\newblock Pruning filters for efficient convnets.
\newblock {\em arXiv preprint arXiv:1608.08710}, 2016.

\bibitem{lin2017focal}
Tsung-Yi Lin, Priya Goyal, Ross Girshick, Kaiming He, and Piotr Doll{\'a}r.
\newblock Focal loss for dense object detection.
\newblock In {\em Proceedings of the IEEE international conference on computer
  vision}, pages 2980--2988, 2017.

\bibitem{lin2014microsoft}
Tsung-Yi Lin, Michael Maire, Serge Belongie, James Hays, Pietro Perona, Deva
  Ramanan, Piotr Doll{\'a}r, and C~Lawrence Zitnick.
\newblock Microsoft coco: Common objects in context.
\newblock In {\em European conference on computer vision}, pages 740--755.
  Springer, 2014.

\bibitem{liu2017learning}
Zhuang Liu, Jianguo Li, Zhiqiang Shen, Gao Huang, Shoumeng Yan, and Changshui
  Zhang.
\newblock Learning efficient convolutional networks through network slimming.
\newblock In {\em Proceedings of the IEEE International Conference on Computer
  Vision}, pages 2736--2744, 2017.

\bibitem{long2015fully}
Jonathan Long, Evan Shelhamer, and Trevor Darrell.
\newblock Fully convolutional networks for semantic segmentation.
\newblock In {\em Proceedings of the IEEE conference on computer vision and
  pattern recognition}, pages 3431--3440, 2015.

\bibitem{ma2018shufflenet}
Ningning Ma, Xiangyu Zhang, Hai-Tao Zheng, and Jian Sun.
\newblock Shufflenet v2: Practical guidelines for efficient cnn architecture
  design.
\newblock In {\em Proceedings of the European conference on computer vision
  (ECCV)}, pages 116--131, 2018.

\bibitem{maaten2008visualizing}
Laurens van~der Maaten and Geoffrey Hinton.
\newblock Visualizing data using t-sne.
\newblock {\em Journal of machine learning research}, 9(Nov):2579--2605, 2008.

\bibitem{muller2019does}
Rafael M{\"u}ller, Simon Kornblith, and Geoffrey Hinton.
\newblock When does label smoothing help?
\newblock In {\em NeurIPS}, 2019.

\bibitem{paszke2019pytorch}
Adam Paszke, Sam Gross, Francisco Massa, Adam Lerer, James Bradbury, Gregory
  Chanan, Trevor Killeen, Zeming Lin, Natalia Gimelshein, Luca Antiga, et~al.
\newblock Pytorch: An imperative style, high-performance deep learning library.
\newblock In {\em Advances in neural information processing systems}, pages
  8026--8037, 2019.

\bibitem{rastegari2016xnor}
Mohammad Rastegari, Vicente Ordonez, Joseph Redmon, and Ali Farhadi.
\newblock Xnor-net: Imagenet classification using binary convolutional neural
  networks.
\newblock In {\em European conference on computer vision}, pages 525--542.
  Springer, 2016.

\bibitem{ren2015faster}
Shaoqing Ren, Kaiming He, Ross Girshick, and Jian Sun.
\newblock Faster r-cnn: Towards real-time object detection with region proposal
  networks.
\newblock In {\em Advances in neural information processing systems}, pages
  91--99, 2015.

\bibitem{romero2014fitnets}
Adriana Romero, Nicolas Ballas, Samira~Ebrahimi Kahou, Antoine Chassang, Carlo
  Gatta, and Yoshua Bengio.
\newblock Fitnets: Hints for thin deep nets.
\newblock {\em arXiv preprint arXiv:1412.6550}, 2014.

\bibitem{sandler2018mobilenetv2}
Mark Sandler, Andrew Howard, Menglong Zhu, Andrey Zhmoginov, and Liang-Chieh
  Chen.
\newblock Mobilenetv2: Inverted residuals and linear bottlenecks.
\newblock In {\em Proceedings of the IEEE conference on computer vision and
  pattern recognition}, pages 4510--4520, 2018.

\bibitem{shen2019meal}
Zhiqiang Shen, Zhankui He, and Xiangyang Xue.
\newblock Meal: Multi-model ensemble via adversarial learning.
\newblock In {\em Proceedings of the AAAI Conference on Artificial
  Intelligence}, volume~33, pages 4886--4893, 2019.

\bibitem{shen2017dsod}
Zhiqiang Shen, Zhuang Liu, Jianguo Li, Yu-Gang Jiang, Yurong Chen, and
  Xiangyang Xue.
\newblock Dsod: Learning deeply supervised object detectors from scratch.
\newblock In {\em Proceedings of the IEEE international conference on computer
  vision}, pages 1919--1927, 2017.

\bibitem{shen2021is}
Zhiqiang Shen, Zechun Liu, Dejia Xu, Zitian Chen, Kwang-Ting Cheng, and Marios
  Savvides.
\newblock Is label smoothing truly incompatible with knowledge distillation: An
  empirical study.
\newblock In {\em International Conference on Learning Representations}, 2021.

\bibitem{szegedy2015going}
Christian Szegedy, Wei Liu, Yangqing Jia, Pierre Sermanet, Scott Reed, Dragomir
  Anguelov, Dumitru Erhan, Vincent Vanhoucke, and Andrew Rabinovich.
\newblock Going deeper with convolutions.
\newblock In {\em Proceedings of the IEEE conference on computer vision and
  pattern recognition}, pages 1--9, 2015.

\bibitem{szegedy2016rethinking}
Christian Szegedy, Vincent Vanhoucke, Sergey Ioffe, Jon Shlens, and Zbigniew
  Wojna.
\newblock Rethinking the inception architecture for computer vision.
\newblock In {\em Proceedings of the IEEE conference on computer vision and
  pattern recognition}, pages 2818--2826, 2016.

\bibitem{tan2019efficientnet}
Mingxing Tan and Quoc Le.
\newblock Efficientnet: Rethinking model scaling for convolutional neural
  networks.
\newblock In {\em International Conference on Machine Learning}, pages
  6105--6114, 2019.

\bibitem{tian2019contrastive}
Yonglong Tian, Dilip Krishnan, and Phillip Isola.
\newblock Contrastive representation distillation.
\newblock In {\em ICLR}, 2020.

\bibitem{touvron2019fixing}
Hugo Touvron, Andrea Vedaldi, Matthijs Douze, and Herv{\'e} J{\'e}gou.
\newblock Fixing the train-test resolution discrepancy.
\newblock In {\em Advances in Neural Information Processing Systems}, pages
  8252--8262, 2019.

\bibitem{WahCUB_200_2011}
C. Wah, S. Branson, P. Welinder, P. Perona, and S. Belongie.
\newblock {The Caltech-UCSD Birds-200-2011 Dataset}.
\newblock Technical report, 2011.

\bibitem{walawalkaronline}
Devesh Walawalkar, Zhiqiang Shen, and Marios Savvides.
\newblock Online ensemble model compression using knowledge distillation.
\newblock In {\em ECCV}, 2020.

\bibitem{wu2019detectron2}
Yuxin Wu, Alexander Kirillov, Francisco Massa, Wan-Yen Lo, and Ross Girshick.
\newblock Detectron2.
\newblock \url{https://github.com/facebookresearch/detectron2}, 2019.

\bibitem{yim2017gift}
Junho Yim, Donggyu Joo, Jihoon Bae, and Junmo Kim.
\newblock A gift from knowledge distillation: Fast optimization, network
  minimization and transfer learning.
\newblock In {\em Proceedings of the IEEE Conference on Computer Vision and
  Pattern Recognition}, pages 4133--4141, 2017.

\bibitem{yosinski2014transferable}
Jason Yosinski, Jeff Clune, Yoshua Bengio, and Hod Lipson.
\newblock How transferable are features in deep neural networks?
\newblock In {\em Advances in neural information processing systems}, pages
  3320--3328, 2014.

\bibitem{yun2019cutmix}
Sangdoo Yun, Dongyoon Han, Seong~Joon Oh, Sanghyuk Chun, Junsuk Choe, and
  Youngjoon Yoo.
\newblock Cutmix: Regularization strategy to train strong classifiers with
  localizable features.
\newblock In {\em Proceedings of the IEEE International Conference on Computer
  Vision}, pages 6023--6032, 2019.

\bibitem{zhang2017mixup}
Hongyi Zhang, Moustapha Cisse, Yann~N Dauphin, and David Lopez-Paz.
\newblock mixup: Beyond empirical risk minimization.
\newblock {\em arXiv preprint arXiv:1710.09412}, 2017.

\bibitem{zhang2020resnest}
Hang Zhang, Chongruo Wu, Zhongyue Zhang, Yi Zhu, Zhi Zhang, Haibin Lin, Yue
  Sun, Tong He, Jonas Mueller, R Manmatha, et~al.
\newblock Resnest: Split-attention networks.
\newblock {\em arXiv preprint arXiv:2004.08955}, 2020.

\bibitem{zhang2018shufflenet}
Xiangyu Zhang, Xinyu Zhou, Mengxiao Lin, and Jian Sun.
\newblock Shufflenet: An extremely efficient convolutional neural network for
  mobile devices.
\newblock In {\em Proceedings of the IEEE conference on computer vision and
  pattern recognition}, pages 6848--6856, 2018.

\bibitem{zhu2016trained}
Chenzhuo Zhu, Song Han, Huizi Mao, and William~J Dally.
\newblock Trained ternary quantization.
\newblock {\em arXiv preprint arXiv:1612.01064}, 2016.

\bibitem{zhu2018knowledge}
Xiatian Zhu, Shaogang Gong, et~al.
\newblock Knowledge distillation by on-the-fly native ensemble.
\newblock In {\em Advances in neural information processing systems}, pages
  7517--7527, 2018.

\bibitem{zoph2016neural}
Barret Zoph and Quoc~V Le.
\newblock Neural architecture search with reinforcement learning.
\newblock {\em arXiv preprint arXiv:1611.01578}, 2016.

\end{thebibliography}
}

\end{document}